\documentclass[letterpaper]{article} 
\usepackage[table]{xcolor}
\usepackage{aaai24}  
\usepackage{times}  
\usepackage{helvet}  
\usepackage{courier}  
\usepackage{subcaption}
\usepackage[hyphens]{url}  
\usepackage{multicol}
\usepackage{graphicx} 
\usepackage{natbib}  
\usepackage{caption} 
\usepackage{algorithm}
\usepackage{algorithmic}
\usepackage{amsmath}

\usepackage[capitalize]{cleveref}
\crefname{section}{Sec.}{Secs.}
\Crefname{section}{Section}{Sections}
\Crefname{table}{Table}{Tables}
\crefname{table}{Tab.}{Tabs.}

\usepackage{todonotes}   
\usepackage{color, colortbl}

\usepackage{adjustbox}
\definecolor{Gray}{gray}{0.9}
\newcolumntype{a}{>{\columncolor{Gray}}c}
\setuptodonotes{inline}    
\usepackage{booktabs}

\newcommand*\ouralgo{{\sc MuScaTeL}}

\newcommand*\keyphrase{{\bf Mu}lti-\textbf{Sca}le \textbf{Te}mporal \textbf{L}earning}
\newcommand*\metanet{{\sc Scorer}} 
\newcommand*\geo{{\sc Geo}} 

\newcommand*\ourexp{{\sc Exp}} 
\newcommand*\mixexp{{\sc MixExp}} 
\newcommand*\inst{{\sc Inst}} 
\newcommand*\instexp{{\sc InstExp}} 
\newcommand*\instmixexp{{\sc InstMixExp}}
\newcommand{\probP}{\text{I\kern-0.15em P}}

\usepackage{newfloat}
\usepackage{listings}
\DeclareCaptionStyle{ruled}{labelfont=normalfont,labelsep=colon,strut=off} 
\floatstyle{ruled}
\newfloat{listing}{tb}{lst}{}
\floatname{listing}{Listing}
\title{Instance-Conditional Timescales of Decay for Non-Stationary Learning}

\author {
    Nishant Jain,
    Pradeep Shenoy
}
\affiliations {
    Google Research India\\
    \{nishantjn, shenoypradeep\}@google.com
}

\begin{document}

\maketitle

\begin{abstract}

Slow concept drift is a ubiquitous, yet under-studied problem in practical machine learning systems. In such settings, although recent data is more indicative of future data, naively prioritizing recent instances runs the risk of losing valuable information from the past. We propose an optimization-driven approach towards balancing instance importance over large training windows. First, we model instance relevance using a mixture of multiple timescales of decay, allowing us to capture rich temporal trends. Second, we learn an auxiliary \textit{scorer model} that recovers the appropriate mixture of timescales as a function of the instance itself. Finally, we propose a nested optimization objective for learning the scorer, by which it maximizes forward transfer for the learned model. Experiments on a large real-world dataset of 39M photos over a 9 year period show upto 15\% relative gains in accuracy compared to other robust learning baselines. We replicate our gains on two collections of real-world datasets for non-stationary learning, and extend our work to continual learning settings where, too, we beat SOTA methods by large margins.

\end{abstract}



\section{Introduction}

\begin{figure}[t]
         \centering
         \includegraphics[width=0.8\linewidth]{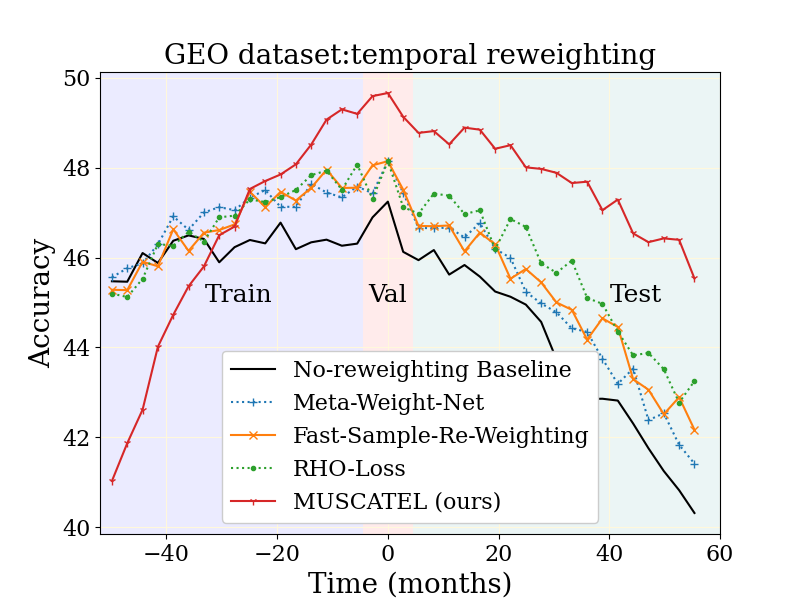}
         \includegraphics[width=.75\linewidth]{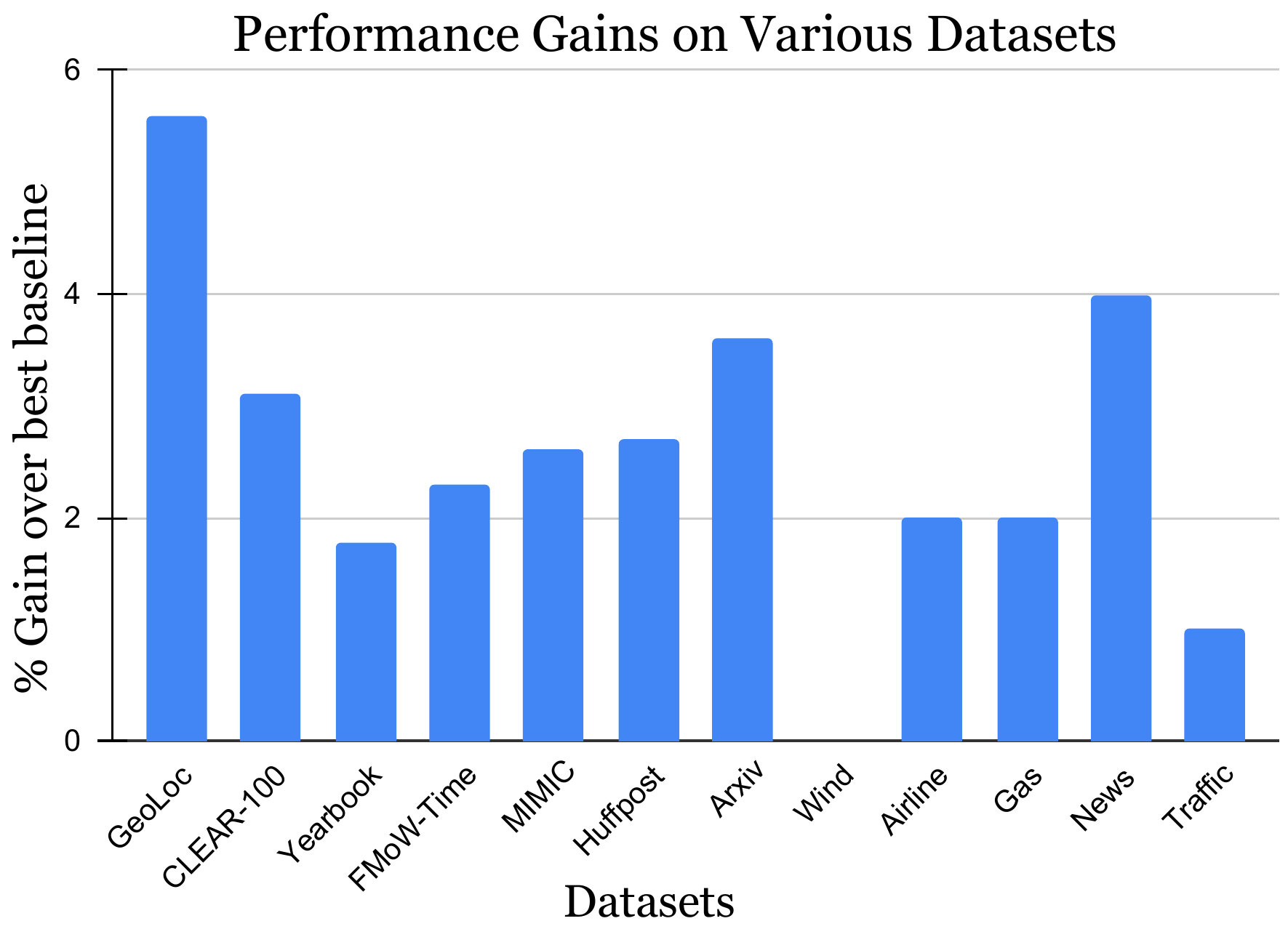}
    \caption{\ouralgo: (a) Our approach significantly improves model performance on future data compared to best-performing baselines. (b) Gains are replicated across a wide range of real-world datasets. See text for details.}
    \label{fig:flowdiag_keyres}
\end{figure}

We study the problem of concept drift--a slow change in patterns of input data and label associations over time--in supervised models trained in an offline or batch-learning fashion (see e.g., \cite{yao2022wild}). Although concept drift is ubiquitous in user-facing AI applications, prevalent practice is still to train batch-learned models from scratch on newer data at regular intervals. By giving equal weight to each training instance, standard batch-learned models overemphasize past, irrelevant data. Alternative approaches favor sequential updates in a streaming data setting, including online learning~\cite{hazan2016introduction}, and continual learning~\cite{zeno2018task,aljundi2019task, Delange_2021}. These approaches are myopic by design (i.e., only have access to the very latest data) and typically overemphasize recent data--in particular, they implicitly downweight past data exponentially as a function of instance age (see e.g., ~\citet{hoeven2018many,jones2023learning}).

We believe that modeling concept drift can benefit substantially from a richer representational language for importance weighting to improve forward transfer, and an optimization-driven approach towards effectively estimating the relevance of past information for future performance. Towards this end, we make 3 technical contributions: First, we model the age-dependent importance of an instance using a mixture of exponential basis functions, allowing significant  flexibility in capturing temporal trends. Second, we customize importance weights for individual instances by learning an auxiliary instance-conditional \textit{importance scoring} model--thus, instance importance is a function of both age and instance properties. Finally, we formulate and efficiently solve a nested optimization problem that jointly learns the supervised model and the scoring model with the mathematical objective of maximizing forward transfer of learned models. 

We call our approach \ouralgo: \keyphrase, and summarize our findings and results below: 

\begin{itemize}
    \item We show substantial gains (\cref{fig:flowdiag_keyres}(a)) on a large, real-world dataset of 39M images over $\sim$ 10 years~\cite{cai2021online}--we improve accuracy and reduce rate of model decay compared to baselines.
    \item We perform a comprehensive evaluation of \ouralgo, combining real-world non-stationary learning datasets from 4 different sources~\cite{cai2021online,yao2022wild,awasthi2023theory,lin2021clear} for a total of 11 batch learning datasets, and 6 continual learning datasets. From an algorithmic perspective, too, we compare against SOTA algorithms in robust learning, meta-learning, and non-stationary learning. \cref{fig:flowdiag_keyres}(b) summarizes our gains on a large number of real-world batch learning datasets over nearest SOTA for each dataset.
    \item  We extend our approach to continual learning settings as well, by performing temporal reweighting within large minibatches, and compare against a range of online and continual learning baselines (see appendix, for an extensive list of baselines). Across datasets and evaluation protocols, \ouralgo\ shows consistent gains over and above all compared algorithms.
    \item We  provide insight into the workings of our algorithm, and in particular our \metanet, showing that it focuses on relevant features and upweights instances in a manner that resonates with visual intuition in addition to being a key component in the substantial accuracy gains. 
\end{itemize}

    

We believe that instance-specific temporal reweighting is a broadly applicable idea for learning under concept drift, and hope that our work contributes to vigorous discussion and development of new algorithms for this important, increasingly relevant problem.

\section{Related Work}\label{sec:relwork}




\subsection{Data Drift and Adaptation}
Early theoretical work on slow concept drift aimed to prove learning bounds under various models of drift such as label drift~\cite{helmbold1994tracking}, and joint distribution drift~\cite{bartlett1992learning}; see also~\citet{barve1996complexity}. Typically they advocated using only recent data for training (a windowed approach), although subsequent work \cite{mohri2012new} introduced a notion of discrepancy that allowed effective use of older data. This discrepancy relies on measuring data-distribution change over time, rather than making overly restrictive assumptions. Recent work \cite{awasthi2023theory} built on this notion and proved more general performance bounds for arbitrary hypothesis function sets. 

On the empirical side, multiple recent papers have proposed benchmark datasets for non-stationary learning, drawing from a wide range of applications (social media, medical records, satellite imagery, etc.) and spanning several years of data collection~\cite{cai2021online,yao2022wild, lin2021clear} (see also~\cite{awasthi2023theory} for additional benchmarks). We merge datasets from these disparate sources to perform a comprehensive, real-world evaluation of our approach applied to both batch-learning and continual learning settings.

\subsection{Learning with Importance Weights}

A number of recent proposals learn instance-specific parameters for training data in order to achieve certain secondary objectives; for instance, improved generalization~\cite{ren2018learning,shu2019meta,mindermann2022prioritized, sivasubramanian2023adaptive}, handling noisy labels~\cite{vyas2020learning}, or implicit curricula for learning~\cite{saxena2019data}. Typically the instance weights are free parameters~\cite{ren2018learning} or functions of instance loss~\cite{shu2019meta}, and not the instance itself. Other work implicitly weights instances by sampling according to some loss criterion~\cite{mindermann2022prioritized}, in order to reduce training cost. 
Finally, some work \cite{sivasubramanian2023adaptive} uses learned reweighting for mixing auxillary losses at an instance-level for improving applications such as distillation.

\subsection{Continual \& Online Learning} 


Continual Learning (CL) and online learning address scenarios where data is available in streaming fashion, and  one instance (or a small buffer of instances) from the stream can be used to update the model before being discarded entirely. Settings include new task acquisition~\cite{van2019three, Delange_2021, de2021continual}, increased output range~\cite{shmelkov2017incremental, Rebuffi_2017_CVPR} or distribution shift whether discrete (domain-incremental CL~\cite{de2021continual, Delange_2021}) or smooth (\textit{concept drift}~\cite{schlimmer1986beyond}). Other work in online learning aims at regret guarantees under certain models of environmental change~\cite{herbster1998tracking}, or combine an adaptive regret objective~\cite{gradu2020adaptive, daniely2015strongly, duchi2011adaptive, hazan2009efficient} with models of environment dynamics.

Due to two key reasons, we focus primarily on the batch learning problem (although we also extend our work to CL settings and compare in them): 1) CL and online learning implicitly downweight past data  in a simplistic manner (exponential downweighting, see e.g.,~\cite{jones2023learning})--this is much less expressive than our instance- and age-dependent model of importance weights, 2) Due to inherent weaknesses (catastrophic forgetting~\cite{mccloskey1989catastrophic}, also related to previous point) and lack of access to past data for iterated learning, CL methods are vastly outperformed by batch learning except in specific settings (see e.g., our comparisons in the appendix).

\section{\ouralgo: Learning with Concept Drift}

We study a learning problem where data is collected over a significantly long period of time, and the data distribution is expected to continually evolve over time. Such concept drift is ubiquitous in practical machine learning systems, and is receiving significant attention in the academic literature recently, in benchmark development~\cite{lin2021clear,yao2022wild} as well as solution approaches~\cite{cai2021online,awasthi2023theory}. We address batch learning in the face of such concept drift, and propose \textit{instance reweighting schemes} that can effectively capture aspects of this temporal evolution, with the mathematical objective of maximizing forward transfer.  

\subsection{Preliminaries}
Consider a supervised data stream with distribution $D_t$ evolving with time, from which samples ($x_t$,$y_t$) are drawn at timesteps $t$. Here, $x_t$ is an input instance with the corresponding label $y_t$. Given data collected upto time $T$ from this stream, we aim to learn a model $f_\theta$: $\mathcal{X} \rightarrow  \mathcal{Y}$ where ($\mathcal{X}$, $\mathcal{Y}$) denote the (input, label) spaces and $\theta$ represents  model parameters, in order to maximize the likelihood of future data upto some time $T+K$: 
\begin{equation}
\prod_{k=\Delta t}^{K}\probP\left(y_{T+k}\left|x_{T+k}, \{x_t, y_t\}_{t=1}^{T}, \theta \right) \right .    
\end{equation}
    
\noindent
The data from $T$ to $T+\Delta t$ is the validation data. To achieve this goal in presence of concept drift, we want to augment the loss on training instances using an importance function $\mathcal{I}{mp}(.)$, such that it more closely represents the loss on future data. This corresponds to decreasing the gap between the expected loss value (for instance cross entropy in classification) for the given and future data:
\begin{equation}
    {E}_{(x,y)\sim D_t}[{E}_{t\sim {p}(t+T)}[l_{(x,y)}] - {E}_{t\sim {p}(t)}[\mathcal{I}{mp}\cdot l_{(x,y)}]]
\end{equation}
where $l_{(x,y)}=l(y, f_\theta(x))$ is the loss function, $p(t)=\frac{1}{N}\sum_{i=1}^N\delta(t_i)$ where $\delta$ is the dirac-delta distribution, and $t_i$ are the ordered timestamps at which any new data-samples were received.
\noindent
We operate in an offline/{batch} setting, where the standard learning strategy is to minimize the expected value of this desired loss over the data distribution $D_t$ via \textit{Empirical Risk Minimization} (ERM):
\begin{equation}
    {E}_{t\sim p(t)}{E}_{(x,y)\sim D_t}[l(f_\theta(x),y)] \approx \frac{1}{T}\sum_{t=1}^{T}{l}(y_t, f_\theta(x_t))
    \label{erm_eq}
\end{equation}

\subsection{Modelling Temporal Drift}\label{sec:reweight}

ERM generally performs well when the data distribution $D$ is static. However, for evolving $D_t$, the approximation in eq. \ref{erm_eq} is poor, leading to high error on test data from the future.  In a real-world scenario, the relationship between $D_{t+\delta t}$ and $D_{1,...,t}$ might be complex and difficult to model in a general manner. Instead, many recent works approximate the dependency using only the most recent data:
\begin{equation}
    \probP(D_{t+\delta t}|D_t, D_{t-\delta t_1}, ....., D_{t-\delta t_n}) \approx  \probP(D_{t+\delta t}|D_t)
\end{equation}
With this myopic view of data evolution, an online learning approach may appear reasonable, as it implicitly places more emphasis on more recent data. In fact, previous work~\cite{hoeven2018many} has drawn an equivalence between online learning and an exponential downweighting of past data. We therefore consider exponential downweighting of data in a batch learning setting as our first baseline. Specifically, instead of equal weights assigned to instances (\cref{erm_eq}), we model importance as as exponentially decaying function resulting in $\hat{p}(t)=p(t)e^{-{\lambda (T-t)}}$ for some fixed $\lambda$ which can be tuned using a validation dataset from the near future. We call this approach \ourexp. This modifies \cref{erm_eq} as follows:
\begin{equation}
    {E}_{t\sim \hat{p}(t)}{E}_{(x,y)\sim D_t}[l(f_\theta(x),y)] \approx \frac{1}{N}\sum_{i=1}^{T}\mathcal{I}{mp}(t){l}(y_t, f_\theta(x_t))
    \label{erm_eq_weighted}
\end{equation}
where the importance $\mathcal{I}{mp}(t) = e^{-\lambda (T-t)}$ is a function of instance age $(T-t)$.  Note, in this proposed approach, we iterate multiple times over all weighted training instances; thus, it has significant advantages over online learning. 

In the above model, using a single exponentially decaying function with a fixed decay rate may limit modeling flexibility. We instead broaden the definition of the importance function $\mathcal{I}{mp}(t)$ to a linear mixture of exponential basis functions:
\begin{equation}
    \mathcal{I}{mp}(t) = \sum_k z_k e^{-a_k(T-t)} = {\mathbf z}^Te^{-{\mathbf a}(T-t)}
    \label{eq:w_calculation}
\end{equation}

This increases the expressivity of $\mathcal{I}{mp}(t)$ by allowing us to model more heavy-tailed functions of time. Here, $\mathbf{a} = \{a_k \}_{k \in \{1,\ldots,K\}}$ are constants, and represent a basis set that capture different timescales of importance decay in the data. The $K$ free parameters $\mathbf{z} = \{z_k \}_{k \in \{1,\ldots,K\}}$ (the mixing weights) assign relative importance to each of the timescales. We call this weighting approach \mixexp.  We make an additional design choice of setting $a_k = a_0^k$ for some fixed $a_0$ -- this allows us to compactly represent a very wide range of timescales, while also reducing the number of free parameters in the formulation. Further, the choices of $a_0, K$ are also not critical, as for moderate $K$, a very wide range of timescales are covered for any choice of $a_0$ ($K=16,a_0=2$ in our experiments), to be mixed by the free parameters $\mathbf{z}$.  Thus, we have $\mathbf{z}$  as the key hyper-parameters to be tuned using the validation set.

\subsection{Instance-Conditional Timescales} 

We come to the final, key component of our proposal for temporal importance weighting: instance-conditional time-scales. In the discussions above, each instance receives an importance weight entirely controlled by its age. In practical settings, however, there are several latent variables that determine the rate of decay of importance. Consider, for instance, topics of discussion on social media--certain topics are reliably constant, while others are more short-lived. This suggests that a one-size-fits-all approach towards temporal reweighting may miss significant opportunities for optimization. To address this, we propose computing the parameters $\mathbf{z}$ as a function of the instance $x$, i.e., 
\begin{equation}
    \mathcal{I}{mp}(x,t) = g(x)^Te^{-{\mathbf a}(T-t)}
    \label{eq:w_calculation_g}
\end{equation}
where the function $g(x)$ now controls the scoring of instance importance in a compact, instance-conditional manner. In particular, we learn an auxiliary neural network (the \metanet) with network parameters $\phi$, i.e., $g_\phi(x)$. Note that we have deliberately chosen to separate instance-specific and time-specific components of $\mathcal{I}{mp}(x,t)$ in \cref{eq:w_calculation_g}; indeed, experimenting with more general functions was not helpful due to the added complexity (see appendix).

\subsection{Learning a \metanet\ for Instance Importance} \label{sec:metanet}
Our goal is to optimize for forward transfer, i.e., the performance of the learned classifier $f_\theta(\cdot)$ on future instances, through the use of the \metanet. This naturally leads to the following objective for the \metanet:
\begin{equation}
     \mathcal{L}^{v} = \frac{1}{M}\sum_{t=T+1}^{T+\Delta t}{l}(y_{t}^v, f_\theta(x^v_t)) 
    \label{eq:val_loss_phi}
\end{equation}
i.e., learn a \metanet\ that minimizes loss on a small amount of data immediately following the training data (denoted as ($x^v_t$, $y^v_t$)). This loss implicitly depends on the primary model's parameters $\theta$, which in turn depends on the \metanet's parameters. More specifically, $\theta$ is learned as:
\begin{equation}
\begin{aligned}
    \theta^*(\phi) = \arg \min_\theta\frac{1}{T}\sum_{t=1}^{T}g_{\phi}(x_t)\cdot e^{-\textbf{a}(T-t)}l(y_t,f_{\theta}(x_t)) \label{eq:outer}
\end{aligned}
\end{equation}
with the choice of $\theta^*$ being a function of the \metanet's parameters $\phi$. Thus, optimizing the \metanet\ objective in \cref{eq:val_loss_phi} can be written as \textit{an outer optimization} over the above optimization for $\theta^*$:
\begin{equation}
 \phi^* =\arg \min_\phi \frac{1}{M}\sum_{t=T+1}^{T+\Delta t}l(y^v_t,f_{\theta^*}(x^v_t))\label{eq:inner}
\end{equation}
where, again, the objective is implicitly a function of $\phi$ through the dependence on $\theta^*(\phi)$. 

\noindent \textbf{Optimization:} The above \textit{bilevel optimization} structure has been used by previous work, e.g., in gradient-based based hyperparameter optimization methods~\cite{lorraine2020optimizing,blondel2021efficient}, and in reweighting for mitigating label noise~\cite{shu2019meta,zhang2021learning}. The former works use implicit gradients for the outer optimization, while the latter used an approximation of a one-step unroll of the inner optimization, and alternating updates to $(\theta,\phi)$. These two approaches have complementary strengths; implicit gradients allow for more precise optimization, while also incurring additional computation costs involving calculating the Hessian. We discuss both below.


 \noindent 
 Given the objective of the \metanet\ in eq. \ref{eq:inner} and $\theta^{\*}$ being a function of $\phi$ (eq. \ref{eq:outer}), the aggregated objective of the \metanet\ can be written as $\mathcal{L}(v) = G^v(\theta^*)$
and its gradient w.r.t. $\phi$ can be calculated using chain rule as follows:
\begin{equation}
    \frac{\partial G^v(\theta^*)}{\partial \phi} = \frac{\partial G^v(\theta^*)}{\partial\theta}\frac{\partial \theta^*}{\partial \phi}
\end{equation}
Calculating the second term in the above equation requires the gradient of optimal classifier parameters with respect to \metanet\ parameters and can be calculated as follows (refer appendix for more details) :

\begin{equation}
    \frac{\partial \theta^*}{\partial \phi} = -\left. \left [ \frac{\partial^2 \mathcal{L}_{tr}}{\partial \theta \partial \theta^{T}} \right ]^{-1} \times  \frac{\partial^2 \mathcal{L}_{tr}}{\partial \theta \partial \phi^{T}} \right|_{\theta^*,\phi}
    \label{hessian}
\end{equation}
where $\mathcal{L}_{tr}$ denotes the training objective from eq. \ref{eq:outer}.
The first term corresponds to the inverse of a Hessian and the second is a second-order term involving gradient w.r.t.  \metanet\ followed by target network parameters. We followed recent work on implicit differentiation to approximate the inverse of the Hessian term~\cite{lorraine2020optimizing}. This leads to a nested optimization setting where we update the \metanet\ for every $L$ updates to the classifier, and the classifier parameters  after these $L$ updates are denoted as $\hat{\theta^*}$, approximating the optimal classifier parameters in \cref{eq:inner}. For more details regarding the implicit gradient and approximation, please refer appendix.

\noindent\textbf{Alternating updates:}
As discussed above,  some works \cite{shu2019meta, ren2018learning}  avoid this term and use an  online approximation to arrive at a single optimization loop with alternating updates to $\theta$ and $\phi$.  We also implemented this variant for our formulation; our findings were qualitatively unchanged, with a modest quantitative trade-off between runtime and accuracy between the two options (see appendix). The final update for $\theta, \phi$, comes out to be as follows:
\begin{equation}
    \begin{aligned}
     \phi_{b+1}= \phi_b + \alpha\beta\frac{1}{MN}\left.\sum_{i=1}^{T}\sum_{j=T+1}^{T+\Delta t}\frac{\partial}{\partial \theta} l(y^v_j,f_\theta(x_j^v)) \right|_{\theta_b} 
     \\ 
    \left.\cdot\frac{\partial}{\partial \phi }g_{\phi}(x_{i_{1}}) \right|_{\phi_b}\cdot e^{-\textbf{a}t_i}\left.\frac{\partial }{\partial \theta} l(y_i,f_\theta(x_{i}))\right|_{\theta_{b}}     \\
    \theta_{b+1} = \theta_{b} - \beta \cdot g_{\phi_{b}}(x_i)\cdot e^{-\textbf{a}t_i}\frac{1}{N}\sum_{i=1}^{T}\left.\frac{\partial }{\partial \theta} l(y_i,f_\theta(x_i))\right|_{\theta_{b}}
    \end{aligned}
\end{equation}

\noindent
where $M=\Delta t$ and $b$ is the number of epochs.
A detailed derivation of the optimization process along with the above approximations is provided in appendix. 



\noindent


\section{Experiment Setup} 


\subsection{Datasets}

\subsubsection{Geolocalization.} 
We experiment extensively with the geolocalization (\geo) dataset proposed by \cite{cai2021online}: 39M images from YFCC100M~\cite{thomee2016yfcc100m}, spanning 2004-2014 and containing natural distribution shifts due to changes in image contents. The task is classification of image region-of-origin (712 geolocations spanning the globe). Images timestamps are used to conduct temporal learning \& evaluation experiments. The authors~\cite{cai2021online} show evidence of gradual distribution shift through a range of experiments in a CL setting.

\subsubsection{Wild-Time Benchmark.}
This collection of 5 datasets captures real-world concept drift~\cite{yao2022wild}--Yearbook, FMoW-time, MIMIC, Huffpost, and arxiv --and spans multiple years. The data covers multiple domains (American high school yearbook photos, satellite images, medical records, news headlines, arxiv preprints). A quick summary is in appendix; please also refer to the source paper for more details. 


\subsubsection{CLEAR Benchmark.} We also experiment with the CLEAR-100 Benchmark~\cite{lin2021clear}, designed to evaluate continual algorithms under realistic distribution shifts and a single-task setting. Like \geo, CLEAR is also sampled from YFCC100M (8M images, sorted in order of timestamps), and aims to capture slow drift in the visual appearance of objects such as laptops, cameras, \textit{etc.}. The authors divide the data into 10 temporal buckets and suggest evaluation schemes for testing CL models.

\subsubsection{Other Real-World drifting datasets.} We further compare on more real-world datasets inheriting concept drift based on a recent paper \cite{awasthi2023theory}. We compare on both regression and classification tasks following the paper. For the regression task Wind, Airline, Gas, News, Traffic datasets are used. We follow the same setup as the \cite{awasthi2023theory} paper, using a completely different source as well as time-segment for testing. The datasets capture the temporal variation ranging from 1 day to 26 years. For more details on these and for description of the classification datasets, please refer appendix. 

\subsection{Batch Training \& Baselines}

Our primary focus is on conventional supervised learning where a model is trained to convergence by iterating over a single, large training dataset. For \geo\ dataset, no baselines for this setup were proposed in the paper, so we propose the following list. Although they were not designed for concept drift, they all (like our approach) use a specialized validation set as ``target'' for optimizing model performance, and indeed do outperform the ERM baseline:\\
\noindent\textbf{MetaWeightNet}: Reweights training instances as a function of instance loss, to minimize loss on a given meta-dataset~\cite{shu2019meta}. \\
\noindent\textbf{Fast sample reweighting}: Loss-based reweighting of instances similar to MetaWeightNet, with automatic selection and updates to the meta-set~\cite{zhang2021learning}. \\ 
\noindent\textbf{RHO-loss}: Ranking training instances for each batch based on minimizing loss on a hold-out set, and selecting top-ranked instances for training~\cite{mindermann2022prioritized}.

\noindent
For the WILD-TIME datasets, we report on all the baselines reported in that paper; likewise, we compare against the already published benchmarks reported in~\citet{awasthi2023theory} for the additional datasets contained there (please refer to the respective papers for details).

We split each dataset into train, validation, and test sets in temporal order, reusing existing partitions where available. Thus,  validation data is more recent than train data. For the \geo\ dataset (see \cref{fig:flowdiag_keyres}(a)), these sets contain 18M, 2M, 19M images covering time periods of around 54,6,60 months respectively. In addition, for this dataset we also held out small data samples from the train and validation period too, so we could compare train/test period accuracy in an unbiased manner.  All methods are trained to converge, and performance on the test period is reported as metric (classification accuracy, or RMSE for regression).  
 

\subsection{Continual Learning \& Baselines}
\label{sec:clbaselines}

We also evaluate our approach in the \textit{offline continual learning setting}, where data is presented to the model in small buckets in sequence, and each bucket is processed and discarded before the next bucket is made available. 

We compare against a suite of recently proposed popular CL baselines including Reservoir Sampling~\cite{kim2020imbalanced}, MiR~\cite{https://doi.org/10.48550/arxiv.1908.04742}, GDUMB~\cite{prabhu2020gdumb}; see appendix for additional comparisons namely ER~\cite{chaudhry2019tiny}, LwF~\cite{li2017learning}, EwC~\cite{kirkpatrick2017overcoming}, AGEM~\cite{chaudhry2018efficient} and other recent methods. In addition, we compare against two recent task free-continual learning methods proposed specifically for non-stationary data streams--Continual Prototype Evaluation\cite{de2021continual} and SVGD \cite{wang2022improving}. Finally, we compare against representative online learning algorithms designed for non-stationary data--Ader~\cite{zhang2018adaptive} and Scream~\cite{zhao2022non}. Please see appendix for detailed descriptions of the various baselines.

We split \geo\ into 39 buckets of 1M images each, with one bucket roughly spanning 3 months. For CLEAR, we use the provided temporal data splits (10 buckets of 160k images each plus a corresponding test set). For adapting \ouralgo\ to this continual setting, we use the most recent 10\%  of each bucket as validation data. Other baselines use the whole bucket (including our validation) as training data. Our temporal reweighting operates only over the time span of the bucket (3 months), since older data are not available in the continual learning setup. All algorithms iterate over each bucket of data $k$ times (k=5 for \geo, and as per the benchmark proposal for CLEAR100). 


\begin{table}[t]
    \centering
    \begin{tabular}{c|ccccc}
    \toprule
         Method $\backslash$ Year  & 1 & 2 & 3 & 4 & 5 \\
           \midrule
         ERM & 45.9& 45.4& 44.1& 42.7& 41.0\\
         MWNet & 46.8& 46.3& 44.8 &  43.7 & 42.0 \\
         FSR & 46.9 & 46.1 & 45.2 & 44.1 & 42.6\\
         RhoLoss & 47.2& 46.7& 46.0& 44.6& 43.3 \\
         \rowcolor{Gray}
         \ouralgo (ours) & 49.1& 48.7& 48.4& 47.6& 47.2\\
         \midrule
        \textbf{\%gain vs ERM }& \textbf{6.7} & \textbf{7.1} & \textbf{9.8} & \textbf{11.5} & \textbf{15}\\
         \textbf{\%gain vs SOTA }& \textbf{3.9} & \textbf{4.1} & \textbf{5.1} & \textbf{6.8} & \textbf{8.8 }\\
         \bottomrule
    \end{tabular}
    \caption{Year-wise error rate for our test set from Geo-localization Dataset. Our method outperform all the other baselines by significant margins.}
    \label{tab:tabularacc_geo}
\end{table}

\begin{table*}[t]
    \centering
    \begin{tabular}{c|cc|cc|cc|cc|cc}
    \toprule
         Method  & \multicolumn{2}{c|}{YB} & \multicolumn{2}{c|}{FMoW} & \multicolumn{2}{c|}{MIMIC} & \multicolumn{2}{c|}{HP} & \multicolumn{2}{c}{Arxiv} \\
         \midrule
   & ID & OOD & ID & OOD & ID & OOD & ID & OOD & ID & OOD \\
           \midrule
         ERM & 97.99 & 79.50& 58.07 & 54.07& 71.30 & 61.33& 79.40 & 70.42 & 53.78 &45.94\\
         GroupDRO-T & 96.04 & 77.06 & 46.57 & 43.87 & 69.70 & 56.12 & 78.04 & 69.53 & 49.78 & 39.06\\
         mixup & 96.42 & 76.72 & 56.93 & 53.67 & 70.08 & 58.82 & 80.15 & 71.18 & 50.72 & 47.82\\
         LISA & 96.56 & 83.65 &55.10 & 52.33 & 70.52 & 56.90  & 78.20 & 69.99  &50.72 & 47.82 \\
         CORAL-T & 98.19& 77.53 & 52.60 & 49.43 & 70.18 &  57.31 & 78.19 & 70.05 & 53.25 & 42.32\\
         IRM-T & 97.02 & 80.46  & 46.60 & 45.00 & 72.33 & 56.53& 78.38& 70.21 &46.30& 35.75\\
         \rowcolor{Gray}
         \ouralgo & \textbf{98.13} & \textbf{82.78} & \textbf{59.87} & \textbf{56.95}& \textbf{72.80} & \textbf{64.12}& \textbf{81.56} & \textbf{73.12} & \textbf{55.89} & \textbf{49.57} \\
         \bottomrule
    \end{tabular}
    \caption{Average accuracy on various datasets of the Wild-Time benchmark for both In-Dist. and OOD due to temporal shift setting. Our method is able to outperform all other methods on both In-Dist. and OOD settings, even though it downweights past. This shows it can robustly learn time-invariant features.}
    \label{tab:tabularacc_wildtime}
\end{table*}

\begin{table}[t]
    \centering
    \begin{adjustbox}{width=0.47\textwidth}
    \begin{tabular}{c|cccccca}
    \toprule
         Method  &KMM & DM & MM & EXP & BSTS & SDRIFT & { Ours}\ \\
           \midrule
         Wind & 1.19 &  1.12 & 1.19 & 0.98 & 0.98 & \textbf{0.95} & \textbf{0.95}\\
         Airline & 2.45 & 1.78 & 1.41 & 0.98 & 0.94 & 0.94 & \textbf{0.92} \\
         Gas  & 0.45 & 0.42 & 0.47 & 0.94 & 1.02 & {0.40} & \textbf{0.38} \\
         News & 1.10 & 1.13 & 1.10 & 0.98 & 1.00 & 0.97 & \textbf{0.93} \\
         Traffic & 2.30 & 2.20 & 0.99 & 0.996 & 0.98 & 0.96 & \textbf{0.95} \\
         \bottomrule
    \end{tabular}
    \end{adjustbox}
    \caption{Comparison with drift datasets and baselines from the SDRIFT paper~\cite{awasthi2023theory}. Our method consistently performs best across all datasets.}
    \label{tab:tabularacc_batchshift}
    \vspace*{-0.2in}
\end{table}

\begin{figure*}[!htb]
    \centering
        \includegraphics[width=.9\linewidth]{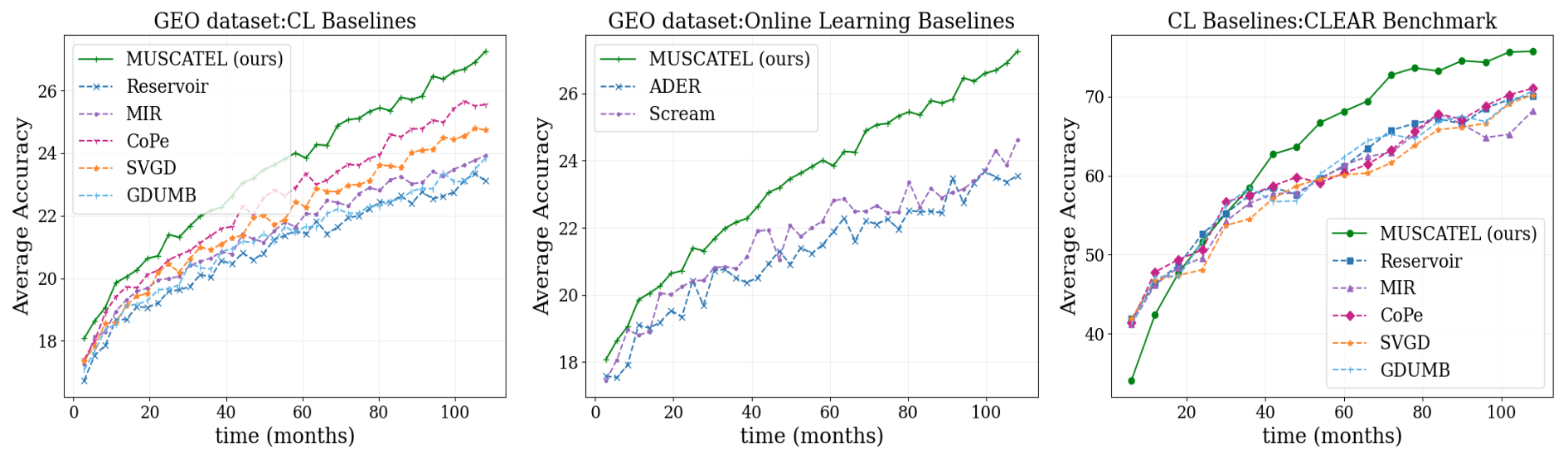}
    \caption{
    Comparison of \ouralgo\ against various continual learning baselines, on the \geo\ dataset, with CL baselines (a) and non-stationary online learning methods (b). Panel (c) shows comparison on the CLEAR-100 concept drift benchmark for continual learning, where, too, we show clear gains.
    } \label{fig:cont}
\end{figure*}

\section{Results}

\begin{figure*}[t]
    \centering
    \includegraphics[width=0.95\linewidth]{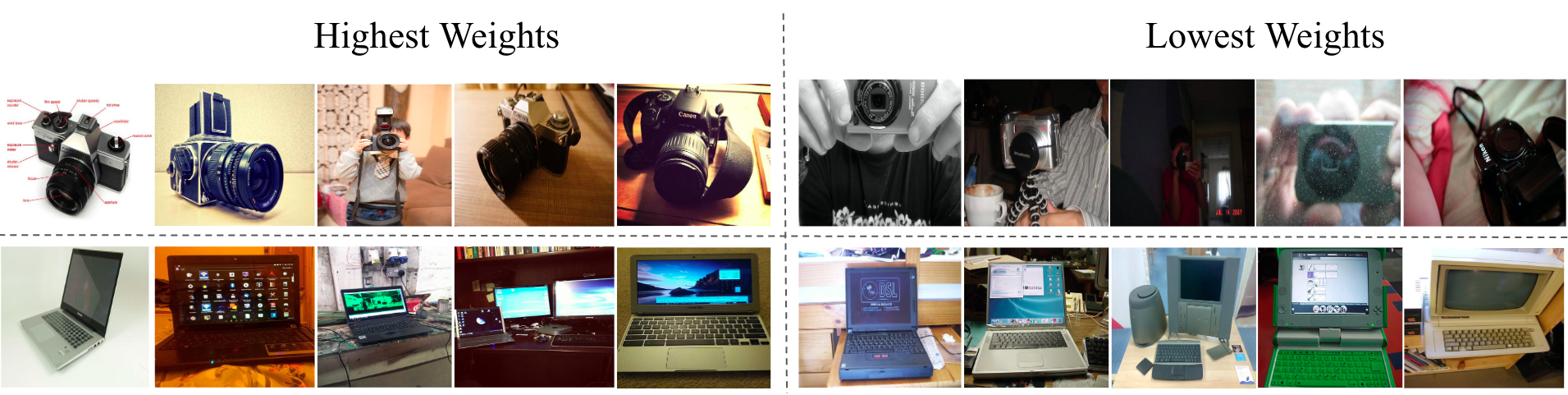}
    \caption{Visualizing the highest weighted examples, using our \metanet\ on the CLEAR-100 benchmark.}
    \label{fig:visual_clear}
\end{figure*}

\subsection{\ouralgo\ Maximizes Forward Transfer}
We first illustrate the complexities of non-stationary learning, and the substantial benefits of our approach, on a large-scale real-world dataset (GEO, 39M photos spanning a 9 year period) (\cref{fig:flowdiag_keyres}(a)), followed by quantitative analysis on a range of benchmark datasets (\cref{tab:tabularacc_geo},\cref{tab:tabularacc_wildtime},\cref{tab:tabularacc_batchshift}).

Figure~\ref{fig:flowdiag_keyres}(a) compares various baselines for batch training on the GEO dataset, over the training period (past data) as well as the testing period (future data). First, ERM has flat performance over the training window, but decays rapidly over time in the test period, confirming the non-iid nature of the data. Most methods beat ERM on the test period, strongly supporting the value of using a \textit{temporally appropriate validation set} as a ``target set'' in concept drift scenarios. \ouralgo\ outperforms other methods by a significant margin over the test period (upto 15\% relative), showcasing the power of our temporal importance weighting approach. Note that \ouralgo\ clearly trades off past accuracy in favor of more recent data; this is due to the meta-objective of maximizing validation accuracy, which is more recent and more representative of future (test) data.

\subsection{Quantitative, Cross-Dataset Gains}
\label{sec:quantbatch}

We first quantified the gains of our approach as a function of time since the models were trained; \cref{tab:tabularacc_geo} compiles these findings (cf. \cref{fig:flowdiag_keyres}(a)). We note a steep decay in performance for all algorithms; however, \ouralgo\ not only beats the baselines consistently but by a widening margin as time passes, with upto 15\% relative gains over ERM, and 8.8\% relative over the nearest baseline.

Next, we replicate our findings on a range of datasets included in the recent Wild-Time concept drift benchmark suite~\cite{yao2022wild}--5 real-world datasets spanning social media, satellite imagery, and medical records over long time spans. Here, older data is used for training and ``in-domain (ID)'' testing, and newer data for ``out-of-domain (OOD)'' testing.  \cref{tab:tabularacc_wildtime} shows model accuracy over training and test periods (past \& future respectively) for a range of baselines included in the benchmark. Note the following trends: a) large gaps between ID \& OOD accuracies, highlighting the data drift, b) No baseline consistently outperforms the others across datasets, and c) \ouralgo\ handily beats all baselines in both in-domain and out-of-domain accuracy. 

We then evaluated another set of non-stationary regression datasets (see e.g.,~\citet{awasthi2023theory} for dataset details \& baselines). Again, \ouralgo\ consistently beats other baselines by noticeable margins in the test period (\cref{tab:tabularacc_batchshift}), confirming the value of our approach in non-stationary offline learning. \cref{fig:flowdiag_keyres}(b) summarizes relative \% gain of our approach across datasets, over and above nearest baseline in each datasets.

\subsection{Adapting \ouralgo\ for Continual Learning}

Although our focus is on importance weighting in the widely-prevalent batch learning scenario, our broader goal is to address learning when data distributions change slowly over time. Since continual learning is an alternate approach for non-stationary learning, we adapted our reweighting scheme for \textit{offline continual learning}, and compared against SOTA techniques in this domain (see Section~\ref{sec:clbaselines} for more details). \ouralgo\ clearly outperforms these baselines on the \geo\ dataset (\cref{fig:cont}a), and many others (see appendix), as well as \textit{online learning} algorithms designed with theoretical guarantees for non-stationary learning (\cref{fig:cont}(b)).  Finally, we also replicated these findings on the recently proposed CLEAR benchmark (\cref{fig:cont}(c)), designed explicitly to test CL algorithms against concept drift.


\subsection{Understanding the \metanet's Role}

We summarize various analyses showcasing the value of our auxiliary \metanet\ model in producing gains, as well as some insights into its workings. Detailed data is presented in the appendix; we only summarize findings here.

\noindent \textbf{Value add:} We compared temporal scoring options on the GEO dataset, and found a strict ordering in performance: ERM $<$ \ourexp $<$ \mixexp $<$ \instmixexp, suggesting that each of the design choices in our approach provided noticeable additional gains. \\
\noindent \textbf{Interpretability:} (1) Post-hoc analysis of \instmixexp\ weights shows a shallower dropoff on average as a function of time compared to exponential fit; however, in any given temporal bucket, the instance weights varied substantially. This supports the idea that the \metanet\ plays a large role in instance-specific customization of weights (cf.~\cref{eq:w_calculation_g}). (2) Figure~\ref{fig:visual_clear} shows the most and least \metanet-weighted images on the CLEAR dataset (a collection of object images through time), clearly showcasing that the \metanet\ emphasizes more modern-looking instances, and conform to our intuitions \& expectations. GradCam analysis~\cite{selvaraju2017grad} of the \metanet's most relevant features confirms that the network focuses on the primary object in the image, and associated relevant features for determining its importance for forward transfer (see appendix).\\
\noindent \textbf{Tuning:} Our results are relatively insensitive to the choice of the parameter $a_0$. Scaling $a_0$ by 0.5,2, or 4 showed minimal influence on the results. Similarly, given the geometric spacing of the $a_i$'s , increasing $K$ beyond a point made no difference. In short, our results are stable over a very broad range of settings for these hyperparameters, by design and in practice (see appendix for details). \\
\noindent \textbf{Cost:} 1) All numbers presented in this work used a simple 4-layer CNN for the \metanet; increasing \metanet\ complexity improved accuracy marginally, up to a point.  2) Per epoch running time for \ouralgo\ is consistently 1.2x-1.5x that of the ERM baseline, and convergence is roughly similar. 

Please see supplementary materials for a range of additional experiments including a more general \metanet\ $g(x,t)$ that is not constrained to a mixture of exponentials, other modifications to the temporal reweighting functions, a head-to-head comparison of CL and batch learning, etc., and implementation details omitted due to lack of space.

\section{Conclusion}
We address the problem of supervised learning in the face of slow concept drift, and propose \ouralgo--instance-specific multi-scale temporal importance weighting--as an approach towards addressing this issue. We propose a \metanet\ to compactly represent instance-specific importance weights, and a nested optimization objective with an efficient implementation for learning the \metanet. Extensive experiments on a suite of real-world datasets confirm substantial concept drift, and show \ouralgo\ providing significant performance gains in both offline/ batch training \& continued training settings. Interestingly, \ouralgo\ improves both backward and forward transfer performance compared to baseline, despite its use of weighted discounting for past data. Our results also show that under nonstationary data conditions, standard CL approaches are at a significant disadvantage compared to batch training, which should be preferred where practicable. 

\bibliography{aaai24.bib}

\appendix

\section{Additional results}
\label{add_results}

\subsection{Batch vs continual learning}
We perform an empirical comparison of batch and continual learning approaches in the context of the \geo\ dataset. Although continual learning has several advantages--low cost of (incremental) model update, and the potential ability to quickly adapt to changes in the data--it also comes with challenges such as \textit{catastrophic forgetting}, wherein learnings from past data are retained poorly. Also, due to the streaming data protocol, one loses fine-grained control in how distant and recent data are optimized together. In contrast, for the batch learning setting, we have access to all past data, and can iterate over all of the data a number of times without risk of overfitting. However, learned models tend to decay quickly, and a new model may need to be trained from scratch at regular intervals. 

\begin{figure}
    \centering
    \includegraphics[width=\linewidth]{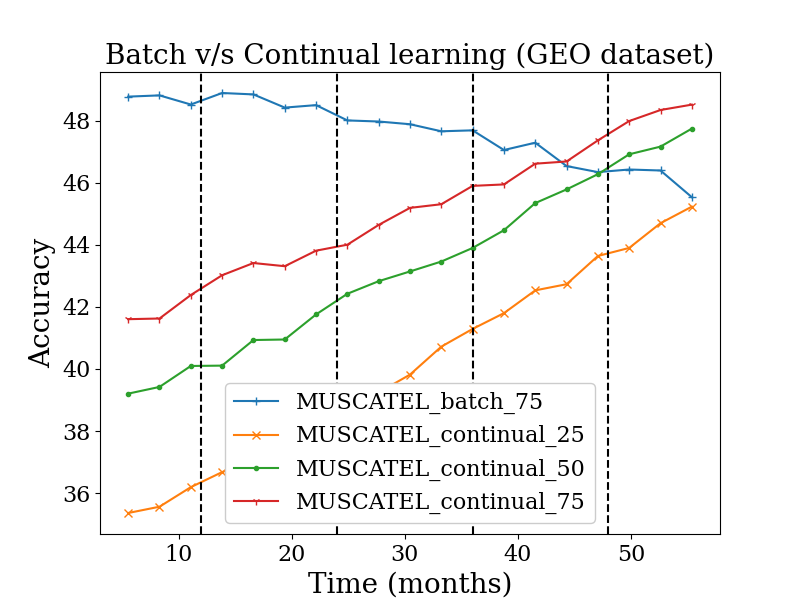}
    \caption{Comparing Batch \& Continual learning in the Geo Dataset. 
  The $x$-axis denotes the time in months with each vertical line corresponding to 1 year.}
    \label{fig:batchvscont}
\end{figure}

Figure~\ref{fig:batchvscont} quantifies these tradeoffs in the context of the \geo\ dataset. The figure corresponds to the test period from Fig 1a, showing a batch-learned \ouralgo\ model trained on 60 months of data, then frozen and evaluated on the following 60 months of test data. To contrast against this, we show continual learning variants of \ouralgo, each updated in sequential manner on data from the smaller time windows (1M data points), and tested on the subsequent window of data. The variants differ in the number of iterations of training performed on each time window. We see the following:
\begin{itemize}
\item A fresh batch-trained model has significantly higher accuracy (around 6\% absolute) than CL variants (t=0 of the testing period), due to the advantages discussed above. However, the batch model decays over time. 
\item CL variants start lower but improve over time due to their continued updates, exposure to more data, and presumably insufficient training. In support of the data paucity theory, the CL methods improve with increased training iterations per chunk (25, 50, 75 iterations); however this gain eventually saturates. 
\item The advantage of the \textit{frozen} batch learning algorithm is retained for a significant time period (around 3.5 years -- each dashed vertical line denotes 1 year of data). This suggests that the modest cost of training a new batch model every 3 years still beats CL alternatives.
\end{itemize}
In summary, depending on the data statistics, rate of change, and the relative costs of model retraining \& deployment vs continual updates, an intermittently retrained batch-learned model may be preferable to a fast-update continual-learned model. However, regardless of the choice of batch versus continual learning for the application, \ouralgo\ can improve performance over vanilla training.

\label{sec:batchvscont}



\subsection{Fine-tuning on warm start models}
\begin{figure}[!t]
    \centering
    \includegraphics[width=\linewidth]{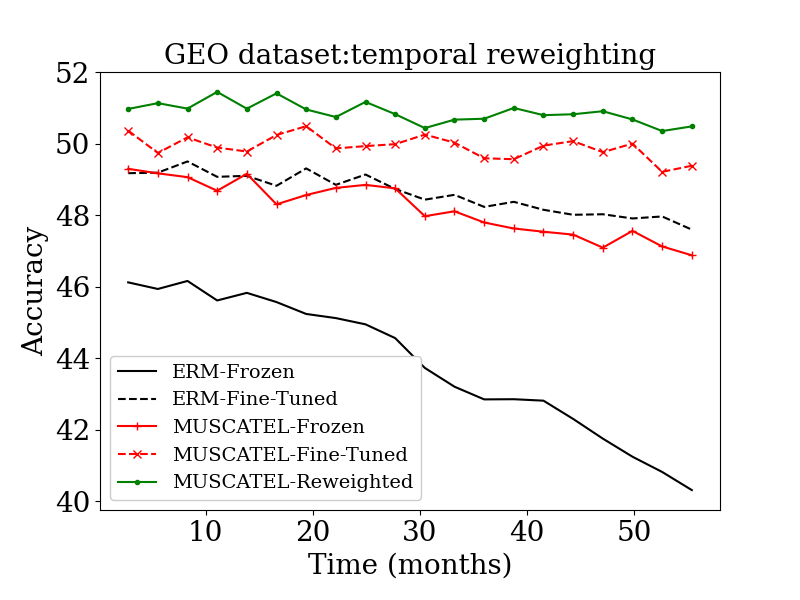}
    \caption{Fine-tuning batch-trained ERM and our model in a windowed fashion on the future data. For reference, we have also provided numbers for their frozen counterparts.}
    \label{fig:warm_start}
\end{figure}

We now analyze running the already trained/ warm-started models in a windowed update setup with naive fine-tuning and with our re-weighting fine-tuning. For this we used a batch-trained ERM model and our \instmixexp\ model. The already trained ERM model is now naively fine-tuned in the windowed fashion, whereas our model is evaluated with both naive fine-tuning and applying our re-weighting scheme again during the windowed fine-tuning. Figure \ref{fig:warm_start} shows the results for this analysis on the future data (test set of their batch training setup). It also contains the results on this future data for the frozen counterparts, \textit{i.e.}, just these given models without any fine-tuning.

 While fine-tuning on warm-started ERM model is competitive with a frozen \ouralgo\ model, fine-tuning on \ouralgo\ base is better. Further, combining \ouralgo's warm start model with \ouralgo's windowed reweighting (our approach in Sec. 4.3) performs even better.\\

\subsection{Comparison with more CL baselines}
\begin{figure*}[th]
    \centering
    \includegraphics[width=0.49\linewidth]{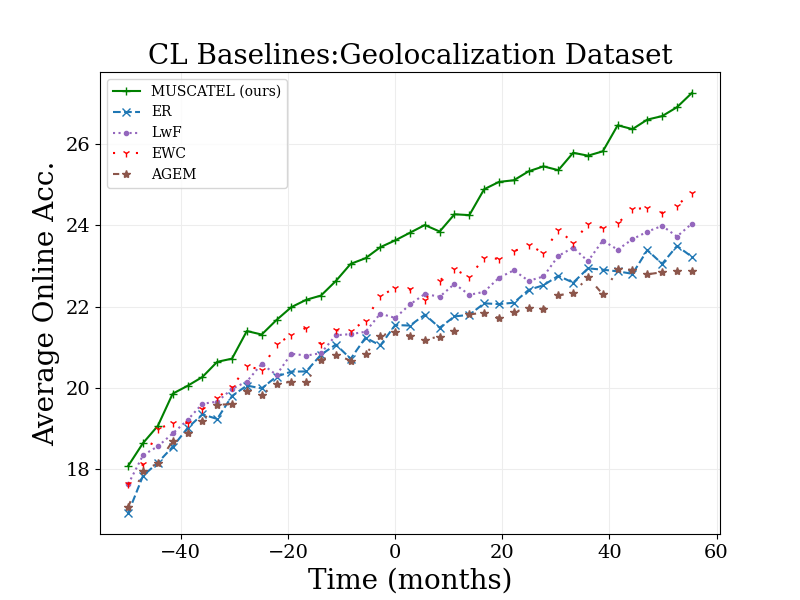}
    \includegraphics[width=0.49\linewidth]{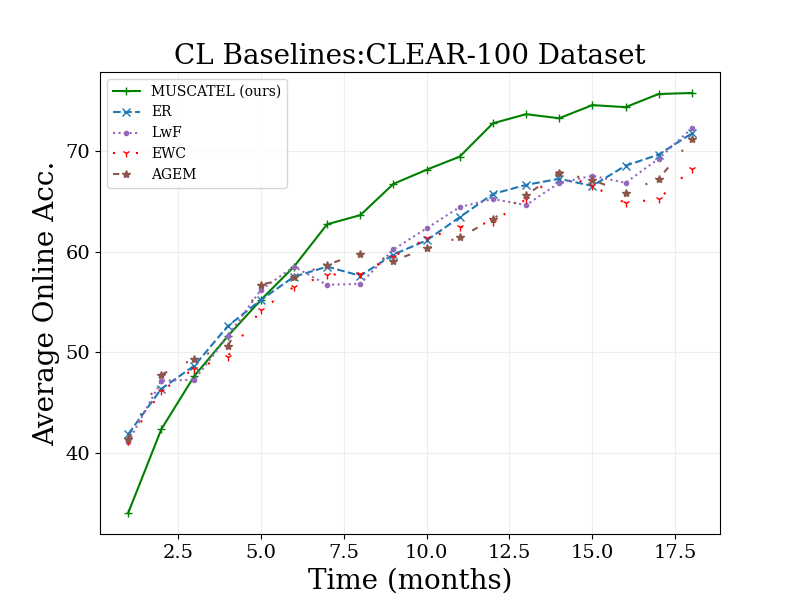}
    \caption{ Comparison of \ouralgo\ (with the \instmixexp\ weighting scheme) against various continual learning baselines, on the geolocalization dataset (left panel) and the CLEAR-100 benchmark (right panel).}
    \label{fig:clbaselines}
\end{figure*}

\noindent
In the paper, we have already compared with popular CL baselines including Reservoir Sampling~\cite{kim2020imbalanced}, MiR~\cite{https://doi.org/10.48550/arxiv.1908.04742}, GDUMB~\cite{prabhu2020gdumb} and also two task-free Continual Learning baselines: Continual Prototype Evaluation\cite{de2021continual} and SVGD \cite{wang2022improving}.
Here, we compare against additional Continual Learning baselines as mentioned in the paper. Specifically, we included the popular methods ER~\cite{chaudhry2019tiny}, LwF~\cite{li2017learning}, EwC~\cite{kirkpatrick2017overcoming}, AGEM~\cite{chaudhry2018efficient}. Figure \ref{fig:clbaselines} shows the results for this comparison, for both geolocalization dataset and CLEAR-100 benchmark, under a similar setting as described in the paper. Similar to the scenario observed with the set of baselines analyzed in the paper, here also, our method surpasses the performance of all the baselines, thereby showing the effectiveness of temporal re-weighting in practical continual learning scenarios with concept drift.

\subsection{Influence of pretraining}

\begin{figure}[!htb]
    \centering
    \includegraphics[width=\linewidth]{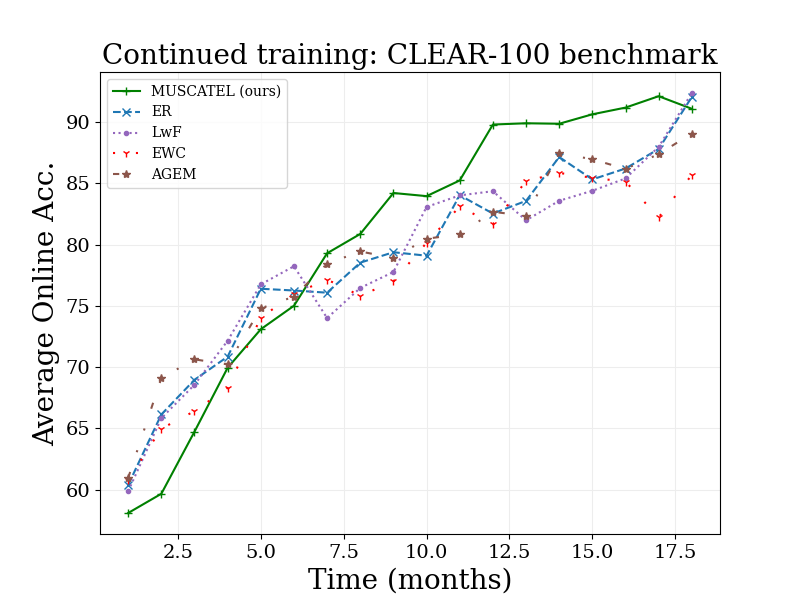}
    \caption{The impact of unsupervised Pretraining on continual learners for the CLEAR-Benchmark.}
    \label{fig:unsup_clear}
\end{figure}

For the CLEAR benchmark,~\cite{lin2021clear}, the authors
show the importance of unsupervised pretraining for learning and adapting models in a continual learning setting. For this purpose, they designate a specific bucket of 0.7M unlabeled image samples prior to the benchmark's time window, and pretrain an unsupervised representation model over this data using  MoCo~V2~\cite{chen2020improved}. The authors show that learning a linear decoder on this pretrained representation can significantly improve accuracy across the entire benchmark time window.  We mimic this evaluation protocol by fine-tuning the linear decoder using \ouralgo\ on top of the pretrained model. Figure \ref{fig:unsup_clear} shows the comparison of our scheme with other standard CL baselines on the extracted features. Even though the performance difference is lower than that observed without any pretraining Fig. 4, as time passes, we see a significant difference in the performance between \ouralgo\ and the other CL baselines.  A point to note is that our \metanet\ in this experiment does not have the advantage of pretraining; the initial lower performance may therefore be explained by a cold-start of the \metanet.  Extending \metanet\ training to exploit unsupervised data is an interesting direction for future work.

\section{Ablations \& extended analyses}

\subsection{Convergence of \ouralgo}
\begin{figure*}[!htb]
        \centering
        \includegraphics[width=0.33\linewidth]{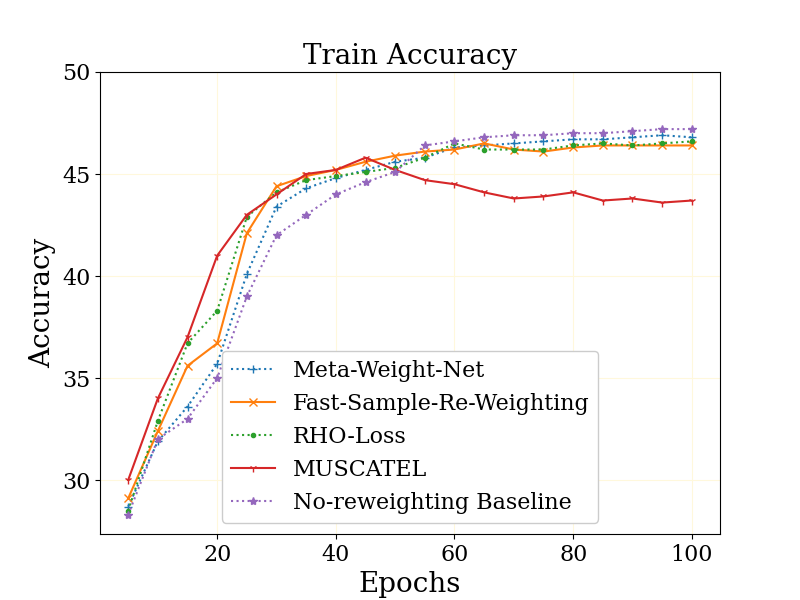}
        \includegraphics[width=0.33\linewidth]{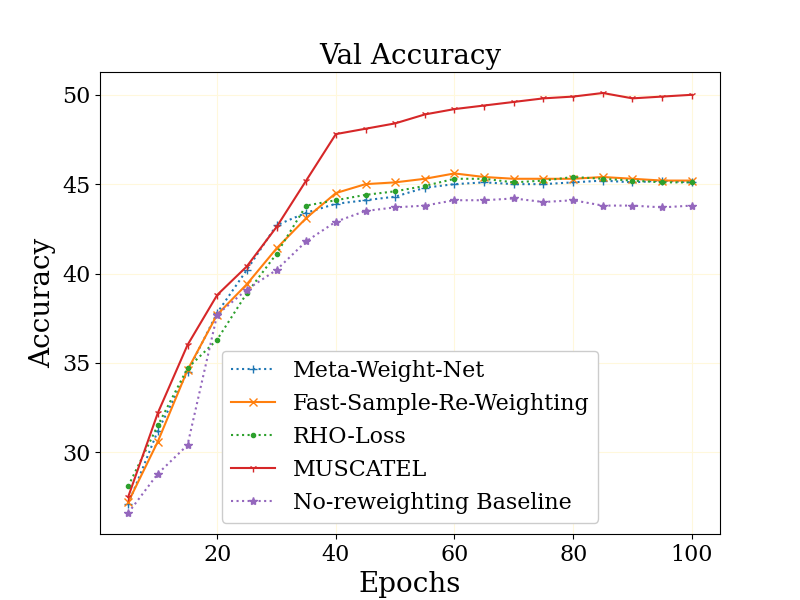}
        \includegraphics[width=0.33\linewidth]{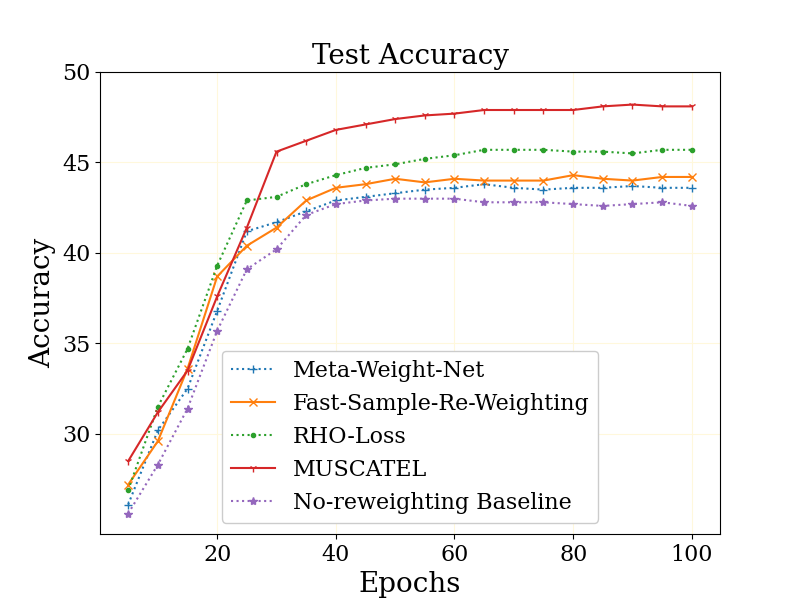}
    \caption{Comparing convergence trends in batch learning. See text for details.}\label{fig:convergence}
\end{figure*}

Figure~\ref{fig:convergence} presents train- validation- and test-window average accuracy as a function of training epochs for \ouralgo\ and the other methods from Fig 1a. We note: a) \ouralgo's training accuracy starts \textit{decreasing} after around 40 epochs--this is the unweighted accuracy, which is trading off older data against newer, b) \ouralgo's validation and test accuracy trends match, and show a continued increase, even after 75 epochs, c)  other algorithms appear to show saturation in validation and test accuracy at an earlier stage, with \ouralgo\ continuing to outperform them. For all methods, a validation-loss-based stopping criterion appears to fairly reflect their performance on the test set.

\subsection{Comparing temporal reweighting methods}
\begin{figure}
    \centering
    \includegraphics[width=\linewidth]{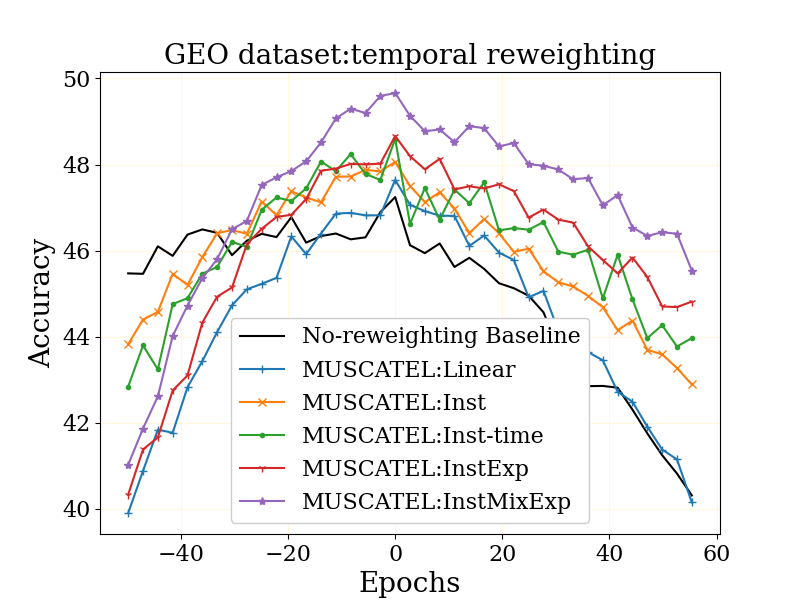}
    \caption{Analysis of the LINEAR, \inst\ , \inst-time, \instexp\ and \instmixexp\ version of \ouralgo\ along with non-reweighting baseline.}
    \label{fig:laters}
\end{figure}

\noindent
We study the contributions of the various components of temporal reweighting to the performance of \ouralgo. Figure~\ref{fig:laters} compares \ouralgo\ with the following reweighting approaches: \instmixexp, \instexp\ (i.e., $K$=$1$), a variant \inst\ which depends only on instance contents and not on time. For \inst, we trained the \metanet\ to directly output a weight given an instance as input, without the use of the timestamp / instance age. We define another variant \inst-time, in which the \metanet\ takes both instance contents and timestamp as input and outputs the weight. Here, the mathematical simplification of separating instance- and temporal contributions is removed ({Eq. 7}). While this indeed provides the \metanet\ with significant additional degrees of freedom in modeling temporal importance, it also removes a key, statistically inspired inductive bias of exponential decay, making learning the function potentially challenging. Finally, we include a linear decay variant which also uses a single parameter as the linear decay rate, to be searched using the validation set.

From the performance comparison, we draw the following conclusions:
\begin{itemize}
    \item \inst\ is better than baseline: this shows that instance-specificity of weights is important for improving performance, even in the absence of temporal information (also see next section).
    \item \inst-time improves on \inst, showing that giving temporal information adds on top of instance-specificity.
    \item \instexp\ is better than \inst-time, suggesting that instead of directly feeding timestamp as the input to the \metanet\ its better to use it separately as an exponentially decaying term.    
    \item \instmixexp\ is best overall, supporting the need for modeling multiple timescales in the data. 
    \item Linear downweighting is somewhat worse to using exponential decay and is significantly worse than \instmixexp\ .
\end{itemize}

In this manner, we see that each of our reweighting design choices provide additive value in \ouralgo. 


\subsection{Varying \metanet\ and classifier architectures}

\begin{figure*}[!htb]
    \centering
    \includegraphics[width=.48\textwidth]{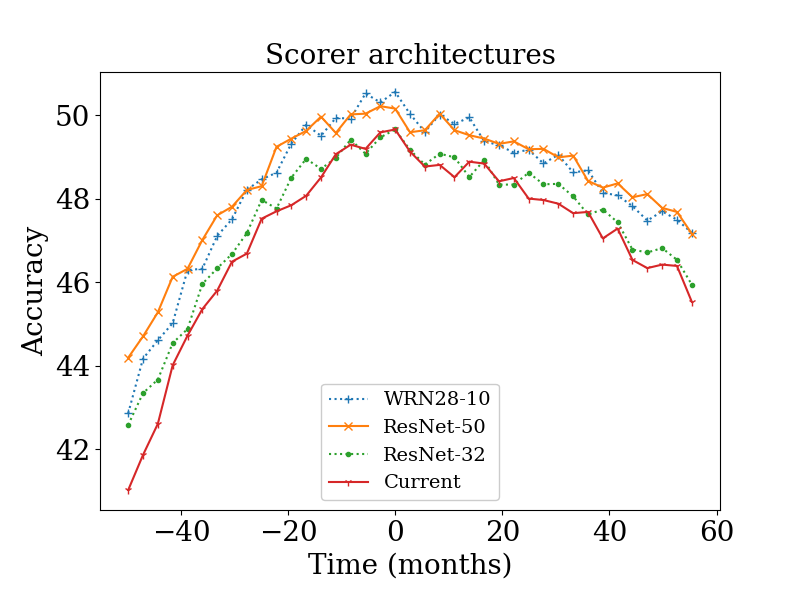}
    \includegraphics[width=0.48\textwidth]{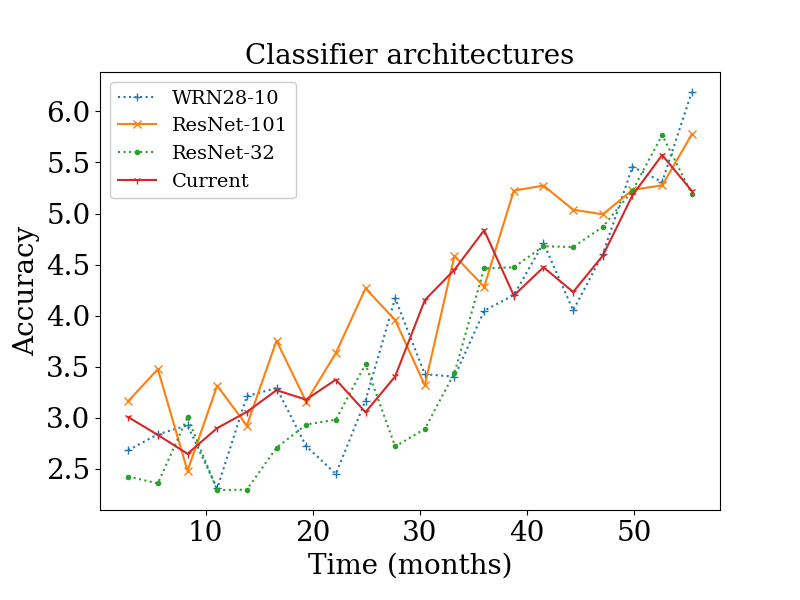}
    \label{fig:arch_abl}
    \caption{Comparing different architectures for classifier (target) network and meta-network with the architecture used in the paper (Current).}
\end{figure*}

\cref{fig:arch_abl} (left) compares 4 \metanet\ architectures on the experiments in {Fig. 1a} (pretrained WRN28-10, ResNet-32, ResNet-50, current). Increased \metanet\ capacity gives modest additional gains. \cref{fig:arch_abl} (right) shows \ouralgo\ gains during test period vs baseline for 4 primary model (i.e., classifier) architectures: WRN28-10, ResNet-32, ResNet-101, Resnet-50. Increased model complexity improves both \ouralgo\ \& baseline (data not shown); however, the relative gains of \ouralgo\ versus baseline remains similar across architectures, and are increasing over the test period since the baseline decays rapidly. 

\section{\metanet: Analysis \& insights}

\subsection{Weight dependence on time}
\begin{figure}[!htb]
    \centering
    \includegraphics[scale=0.5]{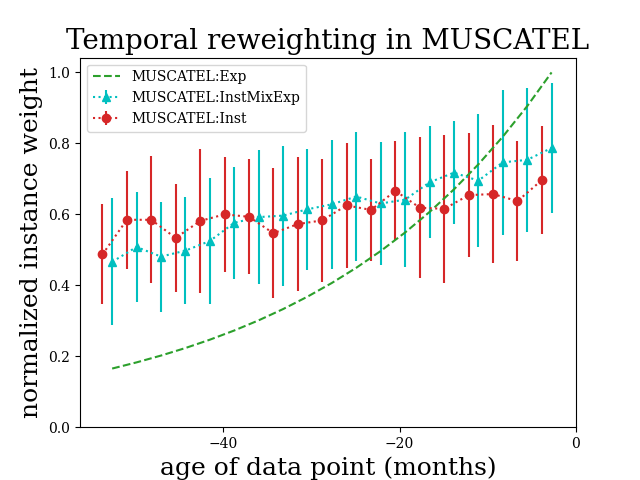}
    \caption{Analysis of variation in average weights of the buckets along with the std w.r.t. time for \ourexp, \inst, \instmixexp. }
    \label{fig:w_time}
\end{figure}

\noindent
We  examine the variation of learned instance weights as a function of time (i.e., instance age) for the following variants of \ouralgo: \inst, \instmixexp\ and \ourexp\ version. We ordered the training data by age and computed averages over instances within short time windows (buckets). We plot the average and standard deviation of weights corresponding to instances within each bucket, against the average instance age of the bucket. \cref{fig:w_time} shows this analysis.

By design, \ourexp\ is a simple exponential reweighting, with less weight on older data and a fixed weight determined by the age of the instance alone. \instmixexp\ on the other hand shows the following interesting characteristics: a) a shallower reduction over time on average--this is through the use of mixture of exponentials, b) wide errorbars on each time window (errorbars = 1 std), indicating the significant variation induced by the instance-conditioning in the \metanet.  From the previous empirical analyses, we've shown that \instmixexp\ does indeed significantly outperform \ourexp, and the variation within-bucket (i.e., the instance-conditioning) is the source of this large gain.

Interestingly, the weights assigned by \inst\ show a modest decay over time (although much less steep compared to \ourexp), even though the \inst\ weighting scheme does not have access to the age of the data point. This shows that the \metanet\ is able to extract some properties of the data instance that partially underlie the concept drift over time,\footnote{There are a number of potential reasons for this drift such as camera/picture quality / resolution of the images, or even some small shifts in label distribution. See subsequent analyses for more insights.} and correctly identify (and downweight) older samples. 

\subsection{Understanding the \metanet's behavior}
\begin{figure*}[t]
    \centering
    \includegraphics[width=0.95\linewidth]{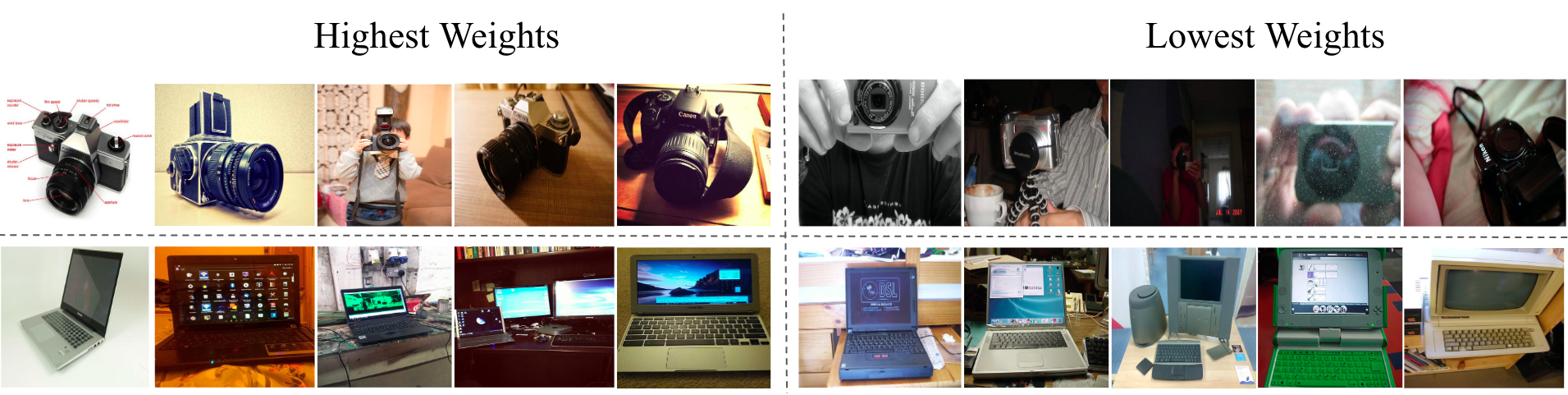}
    \caption{Visualizing the highest weighted examples, using our \metanet\ on the CLEAR-100 benchmark.}
    \label{fig:visual_clear}
\end{figure*}

\begin{figure*}
    \centering
    \includegraphics[width=0.95\linewidth]{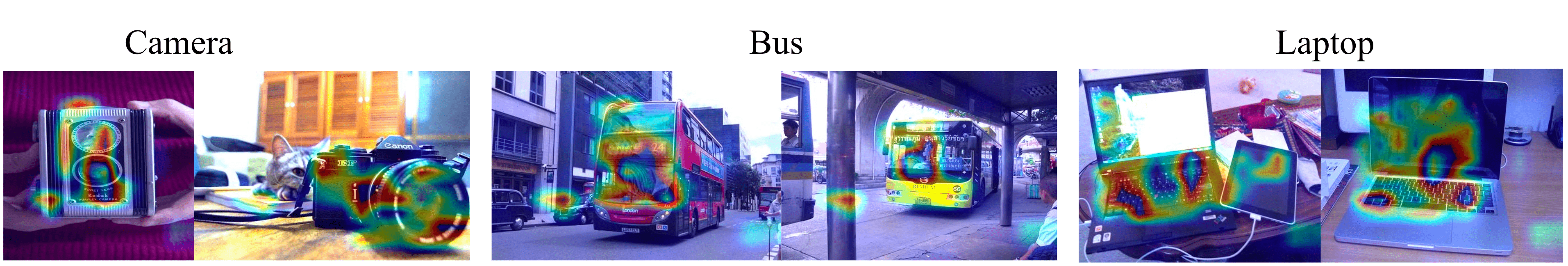}
    \caption{Applying GradCam algorithm to understand the featural basis of the \metanet's working.}
    \label{fig:gradcam_meta}
\end{figure*}

\cref{fig:visual_clear} shows the most and least \metanet-weighted images on the CLEAR dataset (this data is replicated from {Fig. 3}). We see that  \metanet\ emphasizes more modern-looking instances in each object category; this conforms to our visual experience and design goals, that the \metanet\ explicitly optimizes for performance on future data via reweighting training data. Given this qualitative analysis of the weights, we now examine what \metanet\ is actually using from the images in making its decisions. We used the GradCam tool~\cite{selvaraju2017grad}, which generates a pixel heatmap showing the neural network's putative region of interest in processing a particular input. Figure \ref{fig:gradcam_meta} shows this analysis for various objects shown from the CLEAR-100 benchmark; again, the \metanet\ appears to focus primarily on object-related features in calculating importance weights, in keeping with the finding in the previous experiment that object appearance (and related recency) drove the weighting differential across images.

\section{Derivation of update rules}
Here, we further discuss the bilevel optimization scheme we have used in the paper. Revisiting the bi-level optimization objective proposed in the paper:
\begin{equation}
\begin{aligned}
    \phi^* &=\arg \min_\phi \frac{1}{M}\sum_{t=T+1}^{T+M}l(y^v_t,f_{\theta^*}(x^vt)) \\
     \textit{s.t. }  \theta^* &= \arg \min_\theta\frac{1}{N}\sum_{t=1}^{T}g_{\phi}(x_t)\cdot e^{-\textbf{a}(T-t)}l(y_t,f_{\theta}(x_t)) 
\end{aligned}
\label{eq:bilevel}
\end{equation}

where $\theta,\phi$ correspond to model parameters for the primary \& \metanet\ models ($f_\theta$ and $g_\phi$ respectively),  $(x_t,y_t)$ denotes the training instances \& labels\footnote{For ease of notation, $t$ serves both as index over data instances, and the time at which the instance was recorded.},  $(x^v_t, y^v_t)$  the validation instances, and $l$ denotes the loss function. In this nested optimization setting, we first update the parameters of the \metanet ($\phi$) and then using the weights estimated by these updated parameters, we update the model parameters ($\theta$). \\

Let us denote \metanet\ parameters $\phi$, after $e$ epochs on validation data, as $\phi_e$.
By then, there should be $Le$ updates completed on the train data, and the optimal classifier parameters (which we currently approximate as parameters after $L$ updates) should be denoted as $\theta^*_{e}(\phi)$. For updating the \metanet\ a copy of the current classifier is created with parameters dented by $\hat{\theta}$
Revisiting the backpropagation gradient to update $\phi$, from the paper using $\hat{\theta}$, the update equation for the epoch $e+1$ on the validation data is as follows:
\begin{equation}
\begin{aligned}
    \phi_{e+1} &= \phi_e - \alpha\left.\frac{\partial G^v(\hat{\theta}^*)}{\partial \hat{\theta}}\frac{\partial \hat{\theta}^*}{\partial\phi}\right|_{\phi,\hat{\theta}^*} \\
    &= \phi_e - \alpha\left.\frac{\partial G^v(\hat{\theta}^*)}{\partial \hat{\theta}^*}\frac{\partial \hat{\theta}a^*}{\partial\phi}\right|_{\phi=\phi_e,\hat{\theta}^*=\hat{\theta}^*_{e}}
\end{aligned}
\end{equation}
with $G^v(\hat{\theta}^*)$ being the validation loss as a function of copy of current model parameters and $\alpha$ is the learning rate for the \metanet . Further simplifying the equation:
\begin{equation}
\begin{aligned}
    \phi_{e+1} 
     &= \phi_e - \alpha\frac{1}{M}\left.\sum_{t=T+1}^{T+M}\frac{\partial}{\partial \hat{\theta}^*} l(y^v_t,f_{\hat{\theta}^*}(x_t^v)) \right|_{\hat{\theta}^*_{e}}\left. \frac{\partial \hat{\theta}^*}{\partial \phi}\right|_{\phi_e}
\end{aligned}
\label{meta_supp}
\end{equation}
Here, the first gradient term is straightforward to calculate. Let us now see on how to calculate the second term. 

\noindent
\textbf{Calculating $\frac{\partial \hat{\theta}^*}{\partial \phi}$}. Denoting the training objective as $\mathcal{L}_{tr}$ (we assume it denotes the weighted training loss), if there is a stationary point $\hat{\theta}_{\infty}$ (\textit{i.e.} $\left.\frac{\partial\mathcal{L}_{tr}}{\partial \hat{\theta}}\right|_{\phi, \hat{\theta}_\infty}=0$), the following condition holds (Cauchy implicit function theorem):
\begin{equation}
    \frac{\partial}{\partial \phi}\left ( \left. \frac{\partial  \mathcal{L}_{tr}}{\partial \hat{\theta}} \right |_{\hat{\theta}_{\infty},\phi} \right )  = \left ( \frac{\partial^2 \mathcal{L}_{tr}}{\partial \hat{\theta} \partial \phi^T}I + \frac{\partial^2 \mathcal{L}_{tr}}{\partial \hat{\theta} \partial \hat{\theta}^T}\frac{\partial \hat{\theta}_\infty}{\partial \phi} \right ) = 0
\end{equation}
On solving this, the following result can be derived:
\begin{equation}
    \frac{\partial \hat{\theta}_\infty}{\partial \phi} = -\left[ \frac{\partial^2 \mathcal{L}_{tr}}{\partial \hat{\theta} \partial \hat{\theta}^T}\right]^{-1} \frac{\partial^2 \mathcal{L}_{tr}}{\partial \hat{\theta} \partial \phi^T}
    \label{hess}
\end{equation}
We can approximate this stationary point $\hat{\theta}_\infty$ by $\hat{\theta^*}$, by applying $L$ gradient descent updates to classifier. However, due to the time complexity for calculating hessian inverse, this is not a scalable solution for large Neural Nets.
Thus, we follow a recent work  \cite{lorraine2020optimizing} which approximates this inverse hessian term by establishing its equivalence to differentiating the SGD optimization equation of classifier parameters. 

It is based on Neumann series approximation for calculating inverse and assumes a large number of optimization steps for $\theta$.
This results in the following equation:
\begin{equation}
    \frac{\partial \hat{\theta}^*}{\partial \phi} = -\left.\left(\sum_{j<M}\left[ I - \beta\frac{\partial^2 \mathcal{L}_{tr}}{\partial \hat{\theta} \partial \hat{\theta}^T}\right]^{j}\right) \frac{\partial^2 \mathcal{L}_{tr}}{\partial \hat{\theta} \partial \phi^T}\right|_{\hat{\theta}^*}
    \label{hess_approx}
\end{equation}
where $\beta$ denotes the classifier learning rate.  
\\
\noindent
\textbf{Updating Classifier.} Given $Le$ epochs are completed for classifier parameters $\theta$ and the \metanet\ has been updated for $e$ epochs, we denote the classifier parameters as $\theta_{Le}(\phi)$ and its update equation for the next epoch is as follows:
\begin{equation}
    \theta_{Le+1} = \theta_{Le} - \beta \frac{1}{T}\sum_{t=1}^{T}\left.\frac{\partial }{\partial \theta}  g_{\phi_{e}}(x_t)\cdot e^{-\textbf{a}(T-t)} l_{\theta}(x_t,y_t)\right|_{\theta_{Le}}
\end{equation}
Since the parameters $\phi$ are being updated using the model copy with parameters $\hat{\theta}$, the update equation for $\theta$ can be further simplified as:
\begin{equation}
    \theta_{Le+1} = \theta_{Le} - \beta \frac{1}{T}\sum_{t=1}^{T}g_{\phi_{e}}(x_t)\cdot e^{-\textbf{a}(T-t)}\left.\frac{\partial }{\partial \theta}  l_{\theta}(x_t,y_t)\right|_{\theta_{Le}}
\end{equation}

\subsubsection{Alternating Update Scenario.}
Let us again revisit the backpropagation based update equation for $\phi$ for epoch $e+1$, using a copy of model parameters $\hat{\theta}$, but in this alternating optimization scenario it becomes:
\begin{equation}
    \phi_{e+1} = \phi_e - \alpha\left.\frac{\partial G^v(\hat{\theta})}{\partial\phi}\right|_{\phi=\phi_e,\hat{\theta}=\theta_e}
\end{equation}
Further simplifying the equation:
\begin{equation}
\label{eq:supp_phi}
\begin{aligned}
    \phi_{e+1} &= \phi_e - \alpha\left.\frac{\partial G^v(\hat{\theta})}{\partial\hat{\theta}}\right|_{{\theta}_e}\left.\frac{\partial \hat{\theta}}{\partial \phi}\right|_{\phi_e} \\
     &= \phi_e - \alpha\frac{1}{M}\left.\sum_{t=T+1}^{T+M}\frac{\partial}{\partial \hat{\theta}} l(y^v_t,f_{\hat{\theta}}(x_t^v)) \right|_{\hat{\theta}_e}\left. \frac{\partial \hat{\theta}}{\partial \phi}\right|_{\phi_e}
\end{aligned}
\end{equation}
\noindent
Now, again, $\theta$ will be updated using $g_\phi$ and thus, it can be written as $\theta(\phi)$. For epoch $e+1$ it would be updated using $\phi_{e+1}$.
We define $\theta_{e+1}=Q(\theta_{e},\phi_{e+1})$, for some function $Q$, which can further lead to:
\begin{equation}
    \left.\frac{\partial \theta}{\partial \phi}\right|_{\theta_{e+1}, \phi_{e+1}} = \left.\frac{\partial Q(\theta,\phi)}{\partial \phi}\right|_{\theta_{e}, \phi_{e+1}}
\end{equation}
Also, $Q(\theta,\phi)$ represents the following equation:
\begin{equation}
    \theta_{e+1} = \theta_{e} - \beta\frac{1}{T}\sum_{t=1}^{T}\left.\frac{\partial }{\partial \theta}g_{\phi_{e+1}}(x_t)\cdot e^{-\textbf{a}(T-t)} l_{\theta}(x_t)\right|_{\theta_{e}}
    \label{eq:theta_update}
\end{equation}
Differentiating this equation w.r.t. $\phi$ at $e+1$ , we have $\frac{\partial Q(\theta,\phi)}{\partial \phi}$:
\begin{equation}
\begin{aligned}
 \left.-\frac{\beta}{N}\sum_{i=1}^{N}\left.\frac{\partial}{\partial \phi }\right|_{\phi_{e+1}}\frac{\partial }{\partial \theta}g_{\phi_{e+1}}(x_i)\cdot e^{-\textbf{a}(T-t)} l_{\theta}(x_i)\right|_{\theta_{e-1}}     
\end{aligned}
\end{equation}

\noindent
Replacing this in Eq. \ref{eq:supp_phi} by using $\hat{\theta}$, the R.H.S. becomes:
\begin{equation}
    \begin{aligned}
     \phi_{e+1}= \phi_e + \alpha\beta\frac{1}{MT}\left.\sum_{t=1}^{T}\sum_{j=T+1}^{T+M}\frac{\partial}{\partial \hat{\theta}} l(y^v_j,f_{\hat{\theta}}(x_j^v)) \right|_{\theta_e} 
     \\ 
    \left.\cdot\frac{\partial}{\partial \phi }g_{\phi}(x_{i}) \right|_{\phi_b}\cdot e^{-\textbf{a}(T-t)}\left.\frac{\partial }{\partial \theta} l_{\hat{\theta}}(x_{j})\right|_{\theta_{e}}     
    \end{aligned}
\end{equation}

\noindent
Now, for $\theta (\phi)$, from Eq. \ref{eq:theta_update}, the update equation simply reduces to:
\begin{equation}
    \theta_{e+1} = \theta_{e} - \beta \cdot \sum_{i=1}^{T}g_{\phi_{e+1}}(x_i)\cdot e^{-\textbf{a}(T-t)}\frac{1}{N}\left.\frac{\partial }{\partial \theta} l_{\theta}(x_i)\right|_{\theta_{e}}
\end{equation}
since $\phi$ is a function of $\hat{\theta}$ not $\theta$. This involves gradient calculation of only the train loss of instance w.r.t. $\theta$, which is similar to the standard update procedure, and then doing an instance-dependent weighted sum of these gradients.

\section{Implementation \& parameter sensitivity}




\subsection{Baselines}
In the paper, we have compared our methods against several baselines for various datasets. First set of baselines ERM, MWNet \cite{shu2019meta}, RHO-Loss \cite{mindermann2022prioritized}, FSR \cite{zhang2021learning} have already been discussed in the paper.\\
The baselines used for comparison on Wild-Time dataset are as follows:
We have the used the temporally invariant learning baselines alongside the supervised ERM baseline, analysed in the Wild-time dataset paper \cite{yao2022wild}. Following the paper, we have also used the GroupDRO \cite{sagawa2019distributionally}, CORAL \cite{sun2016deep}, IRM \cite{arjovsky2019invariant}, LISA \cite{yao2022improving} and mixup \cite{xu2020adversarial} methods.
The paper modifies the setup for these methods, adapts them to temporal shift setting by dividing the complete time range into various overlapping time-segments (sliding windows) and treating each segment as a domain. We report the numbers for these baseline for In-Distribution (ID) and OOD test sets from the Wild-time \cite{yao2022wild} paper itself. Please refer to the paper for further details regarding their implementation and adaptation to the temporal shift setting. \\
For the datasets used by \citet{awasthi2023theory} in their evaluations, we  used  the baselines analyzed in that paper along with the training methods described there. These include KMM \cite{huang2006correcting}, DM \cite{cortes2014domain}, MM \cite{mohri2012new}, EXP \cite{ross2012exponentially}, BSTS \cite{scott2014predicting} methods. Please refer to the paper for more details on their implementation for the datasets used.\\

\noindent
The baselines used for the Continual Learning setup have already been discussed in the paper and Sec. \ref{add_results} above.

\subsection{Implementation Details}
\noindent
For reproducibility, we provide the training setup along with the hyperparameter details in this section.  The target model architecture is ResNet-50 \cite{he2016deep}, whereas for the \metanet, we use a four-layered CNN architecture with 32 channels each and a max-pooling layer along with ReLU \cite{agarap2018deep} activation and Batch-Normalization layers \cite{ioffe2015batch} in between the network layers, followed by a final fully connected layer. For the target model, we use PoLRS~\cite{cai2021online} in the geolocalization experiments for varying the learning rate across time, initializing with 0.03 for the continued training. In the batched setting we fix the learning to 0.01 for the target model. The \metanet\ was trained with a fixed learning rate of $1e-3$ in all settings.
Also, we use $L=5$ for our optimization scheme. The values for $K$ and $a_k$ are already provided in Sec. \ref{sec:timescale}.\\
For both the datasets discussed in the paper we use a streaming setting where the test set is the stream/bucket at the next discrete timestep. 
For geolocalization dataset we use disjoint buckets each having 1M samples resulting in 39 buckets. For comparison with CL Baselines and Online learning Baselines on the geolocalization dataset, both in the paper and supplementary, 5 epochs per bucker are used. The \metanet\ was trained using the most recent $10$ percent examples from the current data window as validation set for the meta-loss. For the CLEAR-100 benchmark, we use overlapping buckets which means a sliding window with stepsize 7k and total of 18 buckets; all other training and evaluation settings including \# epochs were identical to the setup used in the CLEAR paper. We used the unsupervised-pretrained model provided by the CLEAR-100 Benchmark authors for our pretraining experiment described above.

\subsection{Setting hyperparameters K and $a_k$} 
\label{sec:timescale}
 As mentioned in the paper, $a_k$ are log-linearly spaced; the specific values are $2^2, 2^4, ... 2^{2K}$,  K=8. We chose K using a standard grid search. By design, the spacing of $a_k$ ensure that they span a wide range of values, providing significant coverage in possible timescales of change in the dataset, and that they are scale-invariant, i.e., multiplying all of the $a_k$ by some constant should not matter much, since the \metanet\ can compensate by adjusting the related mixing parameters. As a consequence, we anticipate that simply using the parameters specified above, potentially with a small grid search if desired, should be enough to customize our approach to a new dataset.  

\begin{figure}[!htb]
    \centering
\includegraphics[width=.48\textwidth]{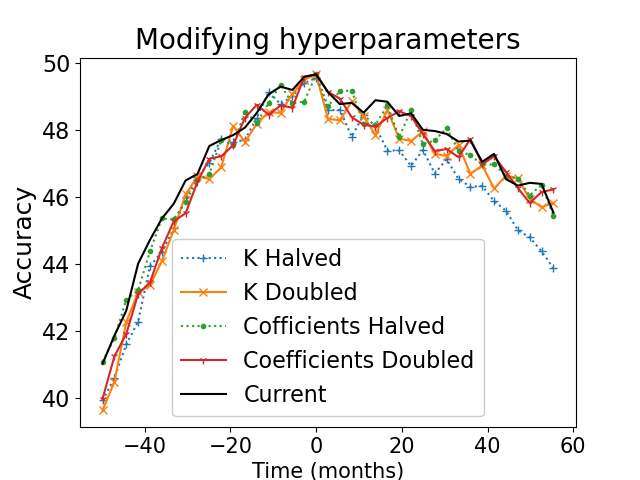}
    \caption{Analyzing different values for \textbf{K} and $a_k$}
    \label{fig:time_abl}
\end{figure}

 These design goals are backed up by empirical evidence-- see Figure~\ref{fig:time_abl} which varies $K$ as well as $a_0$ in our experiment. We see that a) using twice as many timescales $K:=16$ does not improve performance noticeably, and b)scaling $a_k$ by 2, or by 0.5, does not affect results noticeably.
 
\subsection{Application to new datasets}

We used raw timestamps with no additional processing, and we expect the \metanet\ to handle scaling/normalization. For a new dataset, we anticipate that a log-linear scale based on any choice of constant $a_1$, with grid search for $K$, should arrive at a solution that is reasonably robust to perturbation.

\subsection{Implementation of Online Learning baselines}
\noindent
In the paper, we showed analyzed our method against the dynamic regret based online learning baselines which are capable of handling non-stationarity. Specifically we compared it against the Ader\cite{shkodrani2018dynamic} and Scream\cite{zhao2022non} learning algorithms. Given we operate in an offline CL setting and these algorithms are derived/analyzed theoretically on a purely online learning setting, it is not feasible to deploy these in our setting. However, to compare with their instance based dynamic importance weighting schemes with established theoretical bounds for the regret in a changing environment, we deploy them in our setting. Specifically, we apply these re-weighting schemes over batches of data for a given bucket. We then take multiple passes over each of the buckets similar to a offline continued learning setting. \\
Therefore, the complete setting involves initializing the multiple experts and then dynamically weighting them for each batch of data and updating the parameters of the neural network following their update scheme. For each bucket of data there are multiple passes over it, similar to our re-weighting scheme.




\end{document}